# MAA*: A Heuristic Search Algorithm for Solving Decentralized POMDPs

**Daniel Szer** and **François Charpillet**
MAIA Group
INRIA Lorraine, LORIA
54506 Vandœuvre-lès-Nancy, France
{szer,charp}@loria.fr

**Shlomo Zilberstein**
Department of Computer Science
University of Massachusetts
Amherst, MA 01003
shlomo@cs.umass.edu

## Abstract

We present multi-agent A* (MAA*), the first complete and optimal heuristic search algorithm for solving decentralized partially-observable Markov decision problems (DEC-POMDPs) with finite horizon. The algorithm is suitable for computing optimal plans for a cooperative group of agents that operate in a stochastic environment such as multi-robot coordination, network traffic control, or distributed resource allocation. Solving such problems effectively is a major challenge in the area of planning under uncertainty. Our solution is based on a synthesis of classical heuristic search and decentralized control theory. Experimental results show that MAA* has significant advantages. We introduce an anytime variant of MAA* and conclude with a discussion of promising extensions such as an approach to solving infinite horizon problems.

## 1 Introduction

We examine the problem of computing optimal plans for a cooperative group of agents that operate in a stochastic environment. This problem arises frequently in practice in such areas as network traffic routing [Altman 2000], decentralized supply chains [Chen 1999], and control of multiple robots for space exploration [Mataric and Sukhatme 2001]. The general problem has recently been formalized using the Decentralized POMDP (or DEC-POMDP) model [Bernstein et al. 2002]. The model captures effectively the uncertainty about the outcome of actions as well as the uncertainty each agent has about the information available to the other team members. It has been shown that when agents have different partial information about the global state of the environment, the complexity of planning becomes NEXP-complete, which is significantly harder than solving centralized planning problems under uncertainty [Bernstein et al. 2002]. Therefore, finding effective algorithms that provide exact or approximate solutions to this problem is an important challenge. We introduce in this paper the first complete and optimal heuristic search algorithm for solving DEC-POMDPs and evaluate its performance on several test problems.

Heuristic search techniques such as the classical A* algorithm have been shown to be effective for planning when a solution can be constructed incrementally. In general, it means that a set of partial solutions is evaluated, and the most promising one is then selected for further expansion. Bellman's principle of optimality assures such a property for optimal policies in markov decision processes, and A* has thus successfully been applied for planning in MDPs [Hansen and Zilberstein 2001] and POMDPs [Washington 1996]. Heuristic search is particulary beneficial for problems with well-defined initial conditions, such as a single start state distribution. Instead of exploring the entire search space, an appropriate heuristic function may help pruning parts of the space that are not relevant, and thus considerably reduce the amount of computation.

Our main contribution is the application of heuristic search to the domain of multi-agent planning. The difficulty lies in the definition of an evaluation function over a suitable search space. Since agents may have different partial information about the state of the system, it is in general not possible to define state values or belief vectors such as in single agent planning for POMDPs [Hansen et al. 2004]. Furthermore, we know from game theory that there is no optimality criterion for the strategies of a single agent alone: whether a given strategy is better or worse than another strategy depends on the behavior of the remaining agents. Our approach relies on evaluating complete strategy sets, where a strategy set contains a strategy for each agent. We finally describe a method to obtain domain-independent heuristic functions. It has already been shown how the solution of an MDP can be used to efficiently compute an upper bound for a POMDP value function [Hauskrecht 2000]. We exploit a similar property for decentralized domains to compute an optimistic heuristic function over the space of policy sets.

In the remainder of the paper, we first describe recent work on optimal planning for decentralized decision makers, and more specifically introduce the DEC-POMDP model we use. We then propose the multi-agent A* algorithm, present experimental results on three test problems, and discuss some further applications of heuristic search in decentralized POMDPs.

## 2 Related Work

The DEC-POMDP is an extension of the centralized POMDP model which has been shown to have doubly-exponential complexity [Bernstein *et al.* 2002]. Even a two-agent finite-horizon DEC-POMDP is NEXP-complete. Apart from exhaustive search, which becomes quickly intractable when the problem size grows, an optimal algorithm that solves partially observable stochastic games (POSGs) has recently been proposed by [Hansen *et al.* 2004]. It uses dynamic programming (DP) and iterated strategy elimination techniques known from game theory, and is able to treat problems that are otherwise infeasible. However, the algorithm is so far unable to exploit some important additional information such as a fixed and known start state distribution or, in the case of cooperative problems, the presence of a common reward function. Special classes of decentralized POMDPs are sometimes easier to solve, and a first survey has recently been established by [Goldman and Zilberstein 2004]. For problems with independent transitions for example, Becker's coverage set algorithm is guaranteed to find optimal solutions [Becker *et al.* 2004].

Because of the inherent complexity of the DEC-POMDP, several authors have proposed suboptimal algorithms that can treat larger problems. Approximate solutions for POSGs have been proposed by [Emery-Montemerlo *et al.*2004]. A heuristic approach based on Nash's equilibrium theory has been suggested by [Chadès *et al.* 2002] and extended by [Nair *et al.* 2003] in their JESP method, which is guaranteed to converge to an equilibrium solution in the general case. However, the value of such an equilibrium can be arbitrarily bad and depends on the structure of the problem. A multi-agent gradient descent method has been introduced by [Peshkin *et al.* 2000], but it naturally suffers from the common drawback of local optimization algorithms, namely local optimality of the solution. Approximate solutions with a special emphasis on efficient coordination and communication between agents have finally been proposed by [Xuan *et al.* 2001].

## 3 The DEC-POMDP Model

We define the DEC-POMDP model mostly based on [Bernstein *et al.* 2002] although our solution technique is suitable for other decentralized extensions of MDPs such as the MTDP framework defined by [Pynadath and Tambe 2002]. An $n$-agent DEC-POMDP is given as a tuple $\langle S, \{A_i\}, P, R, \{\Omega_i\}, O, T, s_0 \rangle$, where

- $S$ is a finite set of states
- $A_i$ is a finite set of actions, available to agent $i$
- $P(s, a_1, \ldots a_n)$ is a function of transition probabilities
- $R(s, a_1, \ldots a_n)$ is a reward function
- $\Omega_i$ is a finite set of observations for agent $i$
- $O(s, a_1, \ldots a_n, o_1, \ldots o_n)$ is a function of observation probabilities
- $T$ is the problem horizon
- $s_0$ is the start state of the system

Solving a DEC-POMDP for a given horizon and start state can be seen as finding a set of $n$ independent *policies* that yield maximum reward when being executed synchronously. Each agent is assigned a policy that depends only on the local information available to that agent, namely a deterministic mapping from its individual sequence of observations to actions. If we denote $\vec{a}_t = (a_1, \ldots a_n)_t$ as the joint action at time step $t$ of the policy execution, $E[\sum_{t=1}^{T} R(s_t, \vec{a}_t)|s_0]$ represents the expected value we want to maximize.

## 4 Heuristic Search for DEC-POMDPs

A solution to a finite horizon POMDP can be represented as a decision tree, where nodes are labeled with actions and arcs are labeled with observations. Similarly, a solution to a horizon-$t$ DEC-POMDP with known start state can be formulated as a vector of horizon-$t$ policy trees, one for each agent, which are then executed synchronously. We will call such a vector a *policy vector*. Forward search in the space of policy vectors can be seen as an incremental construction of a set of horizon-$(t + 1)$ policy vectors from a parent horizon-$t$ policy vector, which means expanding the leaf nodes of each policy tree in the horizon-$t$ policy vector.

We call $q_i^t$ a depth-$t$ *policy tree* for agent $i$. We denote by $\delta^t = (q_1^t, \ldots q_n^t)$ a *policy vector* of trees, one for each agent, that each have a depth of $t$. We also set $V(s_0, \delta)$ as the *expected value* of executing the vector of policy trees $\delta$ from state $s_0$ (= the expected sum of rewards collected by the agents when executing $\delta$). This value can easily be calculated using the model parameters, and we can state our problem as finding the optimal horizon-$T$ policy vector for a given start state $s_0$:

$$\delta^{*T} = \arg\max_{\delta^T} V(s_0, \delta^T) \quad (1)$$

We build our approach on the popular A* algorithm as a basis for heuristic best-first search. Similarly to A* search, we are able to progressively build a search tree in the space of policy vectors, where nodes at some depth $t$ of the tree constitute partial solutions to our problem, namely policy vectors of horizon $t$. Each iteration of the search process includes evaluating the leaf nodes of the search tree, selecting the node with the highest evaluation, and expanding this node and thus descending one step further in the tree. A section of such a multi-agent search tree is shown in Figure 1.

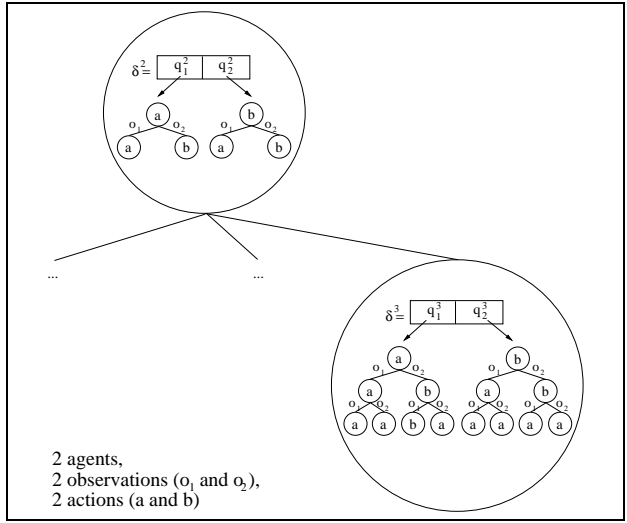

Figure 1: A section of the multi-agent A* search tree, showing a horizon 2 policy vector with one of its expanded horizon 3 child nodes.

Heuristic search is based on the decomposition of the evaluation function into an exact evaluation of a partial solution, and a heuristic estimate of the remaining part. We introduce $\Delta^{T-t}$ as a *completion* of an arbitrary depth-$t$ policy vector, which means a set of depth-$(T-t)$ policy trees that can be attached at the leaf nodes of a policy vector $\delta^t$ such that $\{\delta^t, \Delta^{T-t}\}$ constitutes a complete policy vector of depth $T$. This allows us to decompose the value of any depth-$T$ policy vector into the value of its depth-$t$ root vector ($t \leq T$), and the value of the completion:

$$V(s_0, \{\delta^t, \Delta^{T-t}\}) = V(s_0, \delta^t) + V(\Delta^{T-t}|s_0, \delta^t) \quad (2)$$

The value of a completion obviously depends on the previous execution of the root vector and the underlying state distribution at time $t$. Instead of calculating its value effectively, we are interested in estimating it efficiently. Like in A*, we set

$$F^T(s_0, \delta^t) = V(s_0, \delta^t) + H^{T-t}(s_0, \delta^t) \quad (3)$$

as the *value estimate* for the vector of depth-$t$ policy trees $\delta^t$ and start state $s_0$. The function $H$ must be an *admissible heuristic*, which means it must be an overestimation of the actual expected reward for any completion of policy vector $\delta^t$

$$\forall \Delta^{T-t}: \quad H^{T-t}(s_0, \delta^t) \geq V(\Delta^{T-t}|s_0, \delta^t) \quad (4)$$

In the remaining sections, we will develop multi-agent A* as an extension of classical A* search with policy vectors constituting the nodes of the search tree.

## 5 The Heuristic Function

The core part of the search algorithm remains the definition of an admissible heuristic function. As pointed out by [Littman *et al.* 1995] and later by [Hauskrecht 2000] for the single-agent case, an upper bound for the value function of a POMDP can be obtained through the underlying completely observable MDP. The value function of a POMDP is defined over the space of beliefs, where a belief state $b$ represents a probability distribution over states. The optimal value for a belief state $b$ can then be approximated as follows:

$$V^*_{POMDP}(b) \leq \sum_{s \in S} b(s) V^*_{MDP}(s) \quad (5)$$

Although such belief states cannot be computed in the multi-agent case, a similar upper bound property can still be stated for decentralized POMDPs.

We will now describe a whole class of admissible heuristics and prove that they all overestimate the actual expected reward. In order to do so, we need some more definitions. We set

- $P(s|s_0, \delta)$ as the probability of being in state $s$ after executing the vector of policy trees $\delta$ from $s_0$
- $h^t(s)$ as an optimistic *value function heuristic* for the expected sum of rewards when executing the best vector of depth $t$ policy trees from state $s$

$$h^t(s) \geq V^{*t}(s) \quad (6)$$

with $h^0(s) = 0$

This allows us to define the following class of heuristic functions:

$$H^{T-t}(s_0, \delta^t) = \sum_{s \in S} P(s|s_0, \delta^t) h^{T-t}(s) \qquad (7)$$

Intuitively, any such heuristic is optimistic because it effectively simulates the situation where the real underlying state is revealed to the agents after execution of policy vector $\delta^t$.

**Theorem 5.1.** *Any heuristic function $H$ as defined in (7) is admissible, if $h$ is admissible.*

*Proof.* In order the prove the claim, we need to clarify what happens after execution of policy vector $\delta^t$: each agent $i$ will have executed its policy tree $q_i$ down to some leaf node as a result of a sequence $\theta_i = (o_i^1, \ldots o_i^t)$ of $t$ observations. The completion $\Delta^{T-t}(\theta_i)$ then contains the remaining depth-$(T-t)$ policy tree agent $i$ should execute afterwards. Similarly, the vector $\theta = (\theta_1, \ldots \theta_n)$ represents the individual observation histories for all agents, and the set of depth-$(T-t)$ policy completions for the whole team can thus be written as $\Delta^{T-t}(\theta)$. Its value depends on the underlying state distribution corresponding to the set of observation sequences $\theta$. We can write for the value of any policy completion:

$$V(\Delta^{T-t}|s_0, \delta^t) = \sum_{\theta \in \Theta^t} P(\theta|s_0, \delta^t) V(\Delta^{T-t}(\theta)|s_0, \delta^t)$$

$$= \sum_{\theta \in \Theta^t} P(\theta|s_0, \delta^t) \left[ \sum_{s \in S} P(s|\theta) V(s, \Delta^{T-t}(\theta)) \right]$$

$$\leq \sum_{\theta \in \Theta^t} P(\theta|s_0, \delta^t) \left[ \sum_{s \in S} P(s|\theta) V^{*T-t}(s) \right]$$

$$= \sum_{s \in S} \sum_{\theta \in \Theta^t} P(s|\theta) P(\theta|s_0, \delta^t) V^{*T-t}(s)$$

$$= \sum_{s \in S} P(s|s_0, \delta^t) V^{*T-t}(s)$$

$$\leq \sum_{s \in S} P(s|s_0, \delta^t) h^{T-t}(s) = H^{T-t}(s_0, \delta^t)$$

□

The computation of a good heuristic function is in general much easier than the calculation of the exact value. In our case, the benefit lies in the reduction of the number of evaluations from $|\Omega^{t^n}|$, which is the number of possible observation sequences of length $t$, to $|S|$, the number of states: the value of the heuristic function $h^{T-t}(s)$ has to be calculated only once and should be recorded, since it can be reused for all nodes at level $(T-t)$ as stated in Equation (7). The computation of the value function heuristic $h$ may lead to further savings, and we present several ways to obtain an admissible value function heuristic for our problem.

### 5.1 The MDP Heuristic

An easy way to calculate a value function heuristic is to use the solution of the underlying centralized MDP with remaining finite horizon $T - t$:

$$h^{T-t}(s) = V_{MDP}^{T-t}(s) \qquad (8)$$

Solving an MDP can be done using any DP or search technique and requires only polynomial time. Using the underlying MDP as an admissible heuristic for search in POMDPs has already been applied to the single-agent case as described in [Washington 1996] and later in [Geffner and Bonet 1998]. In our case, the underlying MDP is the centralized and fully observable version of the initial problem, which means it is based on the real system states and joint actions.

### 5.2 The POMDP Heuristic

A tighter yet more complex value function heuristic constitutes in using the solution to the underlying partially observable MDP:

$$h^{T-t}(s) = V_{POMDP}^{T-t}(s) \qquad (9)$$

Although the underlying POMDP is partially observable, it still considers joint actions and thus overestimates the expected sum of rewards for the decentralized system. Solving POMDPs is PSPACE-complete and usually involves linear programming. The fastest algorithm to solve them in general is *incremental pruning* [Cassandra *et al.* 1997] and its variants. However, any upper bound approximation to the exact POMDP solution such as the *fast informed bound* method [Hauskrecht 2000] or any search method can also be used as an admissible heuristic.

### 5.3 Recursive MAA*

The closer the heuristic is to the optimal value function, the more pruning is possible in the search tree. An important special case of a value function heuristic thus constitutes the optimal value itself: $h^t(s) = V^{*t}(s)$. It is the tightest possible heuristic. One way to compute this value efficiently is to apply MAA* again:

$$h^{T-t}(s) = MAA^{*T-t}(s) \qquad (10)$$

This leads to a recursive approach, where a call to $MAA^{*T}$ invokes several calls to $MAA^{*T-1}$, and where each of them launches new subsearches. At each leaf node of the search tree, recursive search thus starts $|S|$ new search subproblems of horizon $T - t$.

In general, a tradeoff exists between the complexity of a tight heuristic function and the possible pruning it allows in the search tree. Our experimental results suggest that recursive MAA* may have some advantages over other heuristic functions, although its computation is more complex (see Section 7). This may be explained by the importance of any additional pruning in an otherwise super-exponentially growing search tree.

## 6 The MAA* Algorithm

We are now able to define the multi-agent heuristic search algorithm MAA*. The root of the search tree is initialized with the complete set of horizon-1 policy vectors, and the search then proceeds similarly as A* by expanding the leaf nodes in best-first order until an optimal horizon-$T$ solution has been identified. Expanding a policy vector $\delta = (q_1, \ldots q_n)$ means: for each leaf node in $q_i$, construct $|\Omega_i|$ child nodes and assign them an action. For a given tree of depth $t$, there are $\Omega^t$ new child nodes and thus $A^{\Omega^t}$ different possible assignments of actions. For a policy vector of size $n$, there will thus be a total number of $(A^{\Omega^t})^n$ new child nodes.

### 6.1 Resolving ties

In classical A* search, nodes are always fully expanded: for a given leaf node, all child nodes are immediatly added to the so called *open list D*. In our case however, this approach presents a major drawback. As pointed out by [Hansen and Zilberstein 1996], suboptimal solutions that are found during the search process can be used to prune the search tree. Since our problem has more than exponential complexity, evaluating the search tree until depth $(T-1)$ is "easy": almost all the computational effort is concentrated in the last level of the tree. This means that suboptimal solutions will be found very early in the search process, and it is thus beneficial not to wait until all leaf nodes of depth $T$ have been evaluated: if a suboptimal solution node with the same value as its parent is found, enumeration of the remaining child nodes can be stoped. Nodes are thus expanded incrementally, which means that only one child assignment of actions is constructed in each iteration step. The parent node remains in the open list as long as it is not fully expanded. Only if the same parent policy vector is selected in a further step, the next possible assignment of actions is evaluated. Together with the tie-breaking property, this may lead to considerable savings in both memory and runtime.

### 6.2 Anytime search

Since solving a DEC-POMDP optimally may require a significant amount of time, it is sometimes preferable to obtain a near optimal solution as quickly as possible, which can then gradually be improved. As already mentioned above, suboptimal solutions are discovered early in the search. It is therefore straightforward to give the algorithm a more *anytime* character by simply reporting any new suboptimal solution that has been evaluated before it is added to the open list.

The discovery of suboptimal solutions in heuristic search can further be sped up through *optimistic weighting*: by introducing a weight parameter $w_i < 1$ on the heuristic, an idea first described by [Pohl 1973], we are able to favor those nodes of the search tree that appear to lead to a suboptimal solution very soon, giving the search a more depth-first character:

$$\overline{F}_i^T(s_0, \delta^t) = V(s_0, \delta^t) + w_i H^{T-t}(s_0, \delta^t) \qquad (11)$$

Althought the convergence criterion remains the $F$-value, expansion of the tree is now guided by the weighted $\overline{F}$-value, which may in addition be time dependant.

### 6.3 MAA*

The generalized MAA* algorithm with incremental node expansion and tie-breaking is summarized in Figure 2. Anytime MAA* can be obtained by replacing the admissible evaluation function $F$ in step 1 of the algorithm with a non-admissible evaluation function $\overline{F}$ as given in (11). We can prove that MAA* indeed returns an optimal solution for any finite horizon DEC-POMDP:

**Theorem 6.1.** *MAA\* is both complete and optimal.*

*Proof.* MAA* will eventually terminate in the worst case after enumerating all possible horizon-$T$ policy vectors, which means after constructing the complete search tree of depth $T$. The leaf node with the highest $F$-value then contains an optimal solution to the problem. If MAA* terminates and returns a policy vector $\delta^T$, the convergence property of A* and the admissibility of the heuristic $H$ guarantees the optimality of the solution. □

MAA* searches in policy space: evaluation and exploration is based on policy vectors. Other evaluation methods within the theory of markov decision processes are based on state values, such that an optimal policy can be extracted by selecting actions that maximize the value of the current state. However, it is not clear if state values can be defined in a similar way for

> Initialize the open list $D_0 = \times_i A_i$
> For any given iteration step $i$:
>
> 1. Select $\delta^* \in D_i$ with $\delta^* \neq \delta^T$ such that $\forall \delta \in D_i$:
>    $F^T(s_0, \delta) \leq F^T(s_0, \delta^*)$
>
> 2. $\delta^{*'} \leftarrow$ Expand $\delta^*$
>
> 3. If $\delta^{*'}$ is an improved suboptimal solution:
>    (a) Output $\delta^{*'}$
>    (b) $\forall \delta \in D_i$:
>        If $F^T(s_0, \delta) \leq F^T(s_0, \delta^{*'}), D_i \leftarrow D_i \setminus \delta$
>
> 4. $D_i \leftarrow D_i \cup \delta^{*'}$
>
> 5. If $\delta^*$ is fully expanded, $D_i \leftarrow D_i \setminus \delta^*$
>
> until $\exists \delta^T \in D_i$ with $\forall \delta \in D_i$:
> $F^T(s_0, \delta) \leq F^T(s_0, \delta^T) = V(s_0, \delta^T)$

Figure 2: The MAA* Algorithm

distributed systems with different partial information. Whether some sort of search in state space is possible for decentralized POMDPs thus remains an important open problem.

## 7  Results

We tested the heuristic search algorithm on three problems, two versions of the multi-agent tiger problem as introduced in [Nair *et al.* 2003], and the multi-access broadcast channel problem from [Hansen *et al.* 2004]. The tiger problem involves 2 agents that listen at two doors. Behind one door lies a hungry tiger, behind the other door are hidden untold riches, but the agents are untold the position of the tiger. Each agent may listen to its door, and thus increase its belief about the position of the tiger, or it may open a door. After a door has been opened, the system is reset to its initial state. In version A of the problem, a high reward is given if the agents jointly open the door with the riches, but a negative reward is given if any agent opens the door of the tiger. Listening incurs a small penalty. In version B, agents are not penalized in the special case where they jointly open the door of the tiger. The multi-agent tiger problem has 2 states, 3 actions and 2 observations.

The multi-access channel problem simulates the transmission of messages at the two ends of a shared network channel. A collision occurs if the agents send a message at the same time. Each agent has no information about the message status of the other agent; however agents get a noisy observation of whether or not a collision occured in the previous time step. The goal of the 2 agents is to maximize the throughput of the channel, and the problem has 4 states, 2 actions, and 2 observations.

| **Problem** | **T=2** | **T=3** | **T=4** |
|---|---|---|---|
| Tiger (A) | -4.00 | 5.19 | 4.80 |
|  | 252 | 105.228 | 944.512.102 |
|  | 8 | 248 | 19.752 |
| Tiger (B) | 20.00 | 30.00 | 40.00 |
|  | 171 | 26.496 | 344.426.508 |
|  | 8 | 168 | 26.488 |
| Channel | 2.00 | 2.99 | 3.89 |
|  | 9 | 1.044 | 33.556.500 |
|  | 3 | 10 | 1.038 |

Figure 3: MAA* using the MDP heuristic (shown are: value of optimal policy, number of evaluated policy pairs, max open list size)

| **Problem** | **T=2** | **T=3** | **T=4** |
|---|---|---|---|
| Tiger (A) | -4.00 | 5.19 | 4.80 |
|  | 252 | 105.066 | 879.601.444 |
|  | 8 | 88 | 18.020 |
| Tiger (B) | 20.00 | 30.00 | 40.00 |
|  | 171 | 26.415 | 344.400.183 |
|  | 8 | 158 | 25.102 |
| Channel | 2.00 | 2.99 | 3.89 |
|  | 9 | 263 | 16.778.260 |
|  | 3 | 6 | 461 |

Figure 4: Recursive MAA* (shown are: value of optimal policy, number of evaluated policy pairs, max open list size)

We tested both the MDP heuristic and the recursive approach on a 3.4GHz machine with 2GB of memory. MAA* with the MDP heuristic explores 33.556.500 policy pairs on the horizon-4 channel problem, which means 3% of the pairs that would have to be evaluated by a brute force approach. However, its actual memory requirements are much less important, because we can delete any suboptimal horizon-$T$ policy pair after evaluating it: the maximum size of the open list thus never exeeds 1.038 policy pairs on that problem. Recursive MAA* performs better on all three test problems, although its heuristic is more complex. This is due to the additional pruning as a result of the tighter heuristic. Runtimes range from a few milliseconds (T=2) to several seconds (T=3) and up to several hours (T=4).

We have compared MAA* to the DP approach from [Hansen *et al.* 2004], to our knowledge the only other non trivial algorithm that optimally solves DEC-POMDPs. Figure 5 shows the total number of policy pairs each algorithm has to evaluate, whereas Figure 6 compares the actual memory requirements. Although heuristic search does not necessarily consider less policy pairs on the sample problem, its memory requirements are much less important. One significant advantage of MAA* over DP thus lies in the fact that it can tackle larger problems where DP will simply run out of memory. For example, MAA* seems to be the first general and optimal algorithm that is able to solve the horizon-4 tiger problems. In addition, it avoids the computationally expensive linear programming part of the DP approach, necessary to identify dominated policy trees. Due to its anytime characteristic, MAA* is furthermore able to return an optimal solution after a very short time, although it might take much longer to guarantee its optimality. On the horizon-4 channel problem for example, the optimal solution is found after a few minutes already. We are equally able to compute suboptimal solutions for larger horizons, although the algorithm will in general run out of time before terminating the computation.

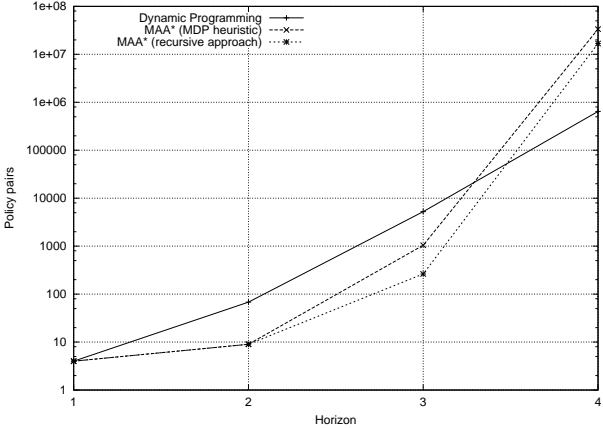
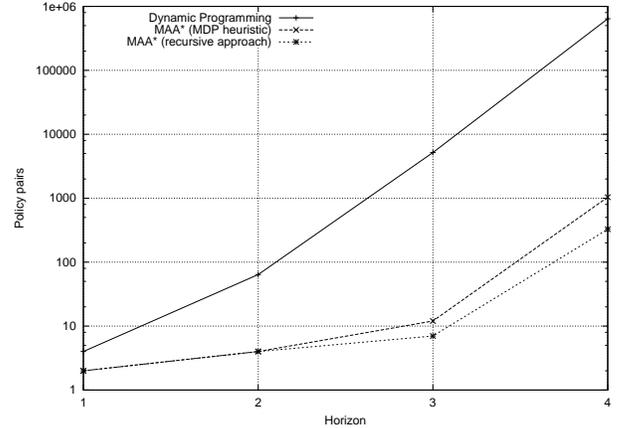

Figure 5: The number of evaluated policy pairs for dynamic programming, MAA* with the MDP heuristic, and recursive MAA* on the channel problem (logarithmic scale).

Figure 6: The actual memory requirements for dynamic programming, MAA* with the MDP heuristic, and recursive MAA* on the channel problem (logarithmic scale).

## 8   Conclusions

We have presented MAA*, the first complete and optimal heuristic search algorithm for decentralized Markov decision processes. Preliminary results show that it compares favorably with the current best solution techniques, and that it presents several advantages with respect to memory requirements and anytime characteristics. MAA* is likely to be one of the current most effective technique for solving finite-horizon DEC-POMDPs and it provides the foundation for developing additional heuristic search techniques for this problem. Although it is so far limited to finite-horizon problems, we are currently investigating extensions to problems with stochastic policies and infinite horizon. The theory of finite state controllers has recently been applied to compute approximate policies for single-agent POMDPs [Hansen 1998], [Meuleau *et al.* 1999], [Poupart and Boutilier 2003]. Generalizing these approaches to multi-agent settings using heuristic search is an important open problem.

## Acknowledgments

We thank Daniel Bernstein and Bruno Scherrer for helpful discussions on this work.